\documentclass[letterpaper]{article} 
\usepackage{aaai24}  
\usepackage{times}  
\usepackage{helvet}  
\usepackage{courier}  
\usepackage[hyphens]{url}  
\usepackage{graphicx} 
\usepackage{natbib}  
\usepackage{caption} 
\usepackage{algorithm}
\usepackage{algorithmic}

\usepackage{newfloat}
\usepackage{listings}
\DeclareCaptionStyle{ruled}{labelfont=normalfont,labelsep=colon,strut=off} 
\floatstyle{ruled}
\newfloat{listing}{tb}{lst}{}
\floatname{listing}{Listing}
\usepackage{multirow}
\usepackage{amssymb}
\usepackage{booktabs}
\usepackage{subcaption}

\title{CoFE-RAG: A Comprehensive Full-chain Evaluation Framework for Retrieval-Augmented Generation with Enhanced Data Diversity }

\author{
    Jintao Liu\textsuperscript{\rm 1},
    Ruixue Ding\textsuperscript{\rm 1,\thanks{Corresponding author}},
    Linhao Zhang\textsuperscript{\rm 2},
    Pengjun Xie\textsuperscript{\rm 1},
    Fie Huang\textsuperscript{\rm 1}
}
\affiliations{
    \textsuperscript{\rm 1}Institute for Intelligent Computing, Alibaba Group\\
    \textsuperscript{\rm 2}University of Chinese Academy of Sciences\\
    \{fengyu.ljt, ada.drx\}@alibaba-inc.com
    
}

\usepackage{bibentry}

\begin{document}

\maketitle
\maketitle

\begin{abstract}
Retrieval-Augmented Generation (RAG) aims to enhance large language models (LLMs) to generate more accurate and reliable answers with the help of the retrieved context from external knowledge sources, thereby reducing the incidence of hallucinations. 
Despite the advancements, evaluating these systems remains a crucial research area due to the following issues:
(1) \textbf{Limited data diversity}: The insufficient diversity of knowledge sources and query types constrains the applicability of RAG systems;
(2) \textbf{Obscure problems location}: Existing evaluation methods have difficulty in locating the stage of the RAG pipeline where problems occur;
(3) \textbf{Unstable retrieval evaluation}: These methods often fail to effectively assess retrieval performance, particularly when the chunking strategy changes.
To tackle these challenges, we propose a \textbf{Co}mprehensive \textbf{F}ull-chain \textbf{E}valuation (CoFE-RAG) framework to facilitate thorough evaluation across the entire RAG pipeline, including chunking, retrieval, reranking, and generation. 
To effectively evaluate the first three phases, we introduce multi-granularity keywords, including coarse-grained and fine-grained keywords, to assess the retrieved context instead of relying on the annotation of golden chunks.
Moreover, we release a holistic benchmark dataset tailored for diverse data scenarios covering a wide range of document formats and query types. 
We demonstrate the utility of the CoFE-RAG framework by conducting experiments to evaluate each stage of RAG systems.
Our evaluation method provides unique insights into the effectiveness of RAG systems in handling diverse data scenarios, offering a more nuanced understanding of their capabilities and limitations.
\end{abstract}

\begin{figure}[t]
\centering
\includegraphics[width=\linewidth]{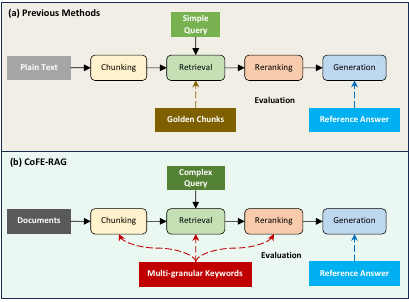} 
\caption{Overview of previous methods and the proposed CoFE-RAG for evaluating RAG systems.}
\label{intro}
\end{figure}

\section{Introduction}
In recent years, Retrieval-Augmented Generation (RAG) has emerged as a powerful paradigm for improving the performance of large language models (LLMs). By integrating the retrieved context with queries, RAG systems can generate more accurate and reliable answers, thereby mitigating the issue of hallucinations that often plagues standalone generative models \cite{DBLP:journals/jmlr/IzacardLLHPSDJRG23}. 
With the development of this technology, comprehensively evaluating all stages of RAG systems becomes increasingly important as it offers guidelines for future improvement and enhances real-world applications.

Mainstream RAG evaluation methods can be broadly divided into reference-free and reference-required methods. Reference-free methods, such as AERS \cite{DBLP:journals/corr/abs-2311-09476} and RAGAS \cite{DBLP:conf/eacl/ESJAS24}, attempt to leverage LLMs to automatically evaluate context relevance, answer relevance, and faithfulness without benchmark datasets. Although these methods bypass the labor-intensive process of data labeling, they suffer from the absence of uniform evaluation standards and the potential risk of introducing subjective bias.
On the other hand, reference-required methods, such as RECALL \cite{DBLP:journals/corr/abs-2311-08147}, RGB \cite{chen2024benchmarking}, and MultiHop-RAG \cite{DBLP:journals/corr/abs-2401-15391},  assess the output of the system against the ground truth reference. 

Despite the promising capabilities of existing RAG evaluation methods, as illustrated in Fig. \ref{intro}, they are still not effective due to the following issues: 
(1) \textbf{Limited data diversity}: The external knowledge base of existing evaluation methods basically derives from well-formed plain text crawled from HTML, which lacks data diversity and makes it difficult to incorporate complex documents such as PDF. Moreover, these methods mainly focus on simple queries, typically factual queries, wherein the answers usually consist of specific entities. This narrows their applicability and hampers their ability to handle more complex analytical or tutorial queries. 
(2) \textbf{Obscure problems location}: Most previous methods predominantly evaluated the end-to-end results without performing step-by-step analysis. The RAG process can be divided into several stages: chunking, retrieval, reranking, and generation. By solely assessing the final generated outcomes, it becomes challenging to identify problems at specific stages within the RAG pipeline. Such approaches would result in poor interpretability and low optimization efficiency, hindering the ability to refine individual components effectively.
(3) \textbf{Unstable retrieval evaluation}: Previous methods evaluate the retrieval stage relying on the annotation of golden chunks with metrics such as Mean Reciprocal Rank and Hit Rate. Annotating all chunks is a tedious and labor-intensive process, and relabeling is required when the chunking strategy is modified. 

To systematically address these challenges, we propose a Comprehensive Full-chain Evaluation (CoFE-RAG) framework to facilitate thorough evaluation across the entire RAG pipeline. We introduce multi-granularity keywords to effectively assess the chunking, retrieval, and reranking phases of RAG systems, which aims to address the dependency on golden chunk annotations for evaluation. The multi-granularity keywords encompass coarse-grained and fine-grained keywords. Specifically, coarse-grained keywords are the most representative and relevant words extracted from the query and context, serving as initial indicators for chunk relevance. Fine-grained keywords are formulated as a set of lists, where each list corresponds to an information point extracted from the context, providing detailed references for answering the query.
CoFE-RAG employs coarse-grained keywords for the initial filtering of retrieved chunks and then uses fine-grained keywords to score the filtered results.

We also release a holistic benchmark dataset specifically designed for diverse data scenarios and can be used to evaluate all stages of RAG systems. This dataset is equipped with a knowledge base encompassing a wide range of document forms. Each example is annotated with the query, multi-granularity keywords, and reference answer. 
We define four types of queries, including factual, analytical, comparative, and tutorial queries.
In order to balance annotation efficiency and annotation quality, we use a combination of LLM automatic annotation and manual review to annotate data. 

In our experimental evaluation, we conduct experiments with various models for each stage of the RAG system to assess their strengths and weaknesses.
The experimental results demonstrate that existing retrieval models excel in handling factual queries but struggle significantly with  analytical, comparative, and tutorial queries.
Furthermore, existing LLMs also perform poorly in leveraging the retrieved context to produce more accurate and reliable responses.
This analysis not only demonstrates the utility of our proposed benchmark but also provides crucial insights on how to optimize each stage of the RAG system.

The main contributions of this paper can be summarized as follows:
\begin{itemize}
    \item We propose the CoFE-RAG framework. To the best of our knowledge, this is the first work to comprehensively evaluate all stages of RAG systems and utilize multi-granularity keywords to improve the evaluation of retrieval results.
    \item This paper releases a benchmark dataset containing four types of queries, multi-granular keywords, and reference answers, along with a knowledge base covering various document formats to evaluate RAG systems in diverse data scenarios.
    \item We conduct a series of experiments to benchmark existing methods at each stage of RAG systems, which facilitates an in-depth analysis of the performance of current RAG systems. The dataset and code are publicly available at \url{https://github.com/Alibaba-NLP/CoFE-RAG}.
\end{itemize}

\section{Related Work}
\subsection{Retrieval-Augmented Generation}
Retrieval-Augmented Generation (RAG) is a technology that combines information retrieval and text generation. It enables LLMs to incorporate retrieved context along with the query to generate more accurate and credible responses, thus reducing the generation of hallucinations \cite{DBLP:journals/jmlr/IzacardLLHPSDJRG23}. 
\citet{DBLP:journals/corr/abs-2301-12652}, \citet{DBLP:conf/acl/YuXY023}, and \citet{DBLP:conf/acl/GaoMLC23} have explored various methods to enhance the effectiveness of retrieval mechanisms. 
\citet{DBLP:conf/iclr/0002IWXJ000023} and \citet{DBLP:journals/corr/abs-2403-00801} investigated the potential for LLMs to directly generate context, effectively bypassing the need for a separate retriever.
\citet{DBLP:journals/corr/abs-2402-10612}, \citet{DBLP:conf/emnlp/WangLSL23}, and \citet{DBLP:journals/corr/abs-2403-14403} used adaptive methods to dynamically determine whether retrieval is necessary to answer a query.
\citet{DBLP:journals/corr/abs-2310-01558}, \citet{DBLP:conf/acl/LiRZWLVYK23}, and \citet{DBLP:journals/corr/abs-2402-18150} aim to enhance the robustness of RAG models.
\citet{DBLP:conf/emnlp/JiangXGSLDYCN23}, \citet{DBLP:journals/corr/abs-2310-11511}, and \citet{DBLP:journals/corr/abs-2403-06840} focused on optimizing the overall RAG pipeline.

\subsection{Retrieval-Augmented Generation Evaluation}
Evaluating the performance of RAG systems has garnered widespread attention, which enables a deeper understanding of the capabilities and limitations of RAG systems.
Evaluation methods for RAG systems can be divided into two main categories: reference-free and reference-required methods.
Reference-free methods, represented by AERS \cite{DBLP:journals/corr/abs-2311-09476} and RAGAS \cite{DBLP:conf/eacl/ESJAS24}, use LLMs to automatically evaluate context relevance, answer faithfulness, and answer relevance without relying on benchmark datasets. 
On the other hand, reference-required evaluations utilize ground truth references to assess the retrieval or generation process, remaining the predominant method for evaluating RAG systems.
For instance, RGB \cite{chen2024benchmarking} aims to evaluate noise robustness, negative rejection, information integration, and counterfactual robustness abilities of LLMs.
RECALL \cite{DBLP:journals/corr/abs-2311-08147} construct a benchmark to evaluate the ability of LLMs to discern the reliability of external knowledge.
CRUD-RAG \cite{DBLP:journals/corr/abs-2401-17043} constructs a large-scale and more comprehensive benchmark to evaluate RAG applications in four distinct tasks: create, read, update, and delete.
MultiHop-RAG \cite{DBLP:journals/corr/abs-2401-15391} propose a comprehensive dataset for evaluating multi-hop queries using a knowledge base derived from news article.
However, these methods fail to provide a comprehensive full-chain evaluation of RAG systems and suffer from limited data diversity.

\begin{table}[t]
	\centering
   \small
	\begin{tabular}{l ccc}
            \toprule
  Format&Avg. Tokens&Avg. Pages&Count\\
  \midrule
  PDF&88495.9&115.4&485\\
  PPT&5662.6&25.9&269\\
  DOC&7894.3&20.2&433\\
  XLSX&3565.2&3.2&227\\
  \midrule
  Total&-&-&1414 \\
   \bottomrule
	 \end{tabular}
	\caption{Distributions of documents in different formats.}
	\label{doc_distribution}
\end{table}

\begin{table}[t]
	\centering
        \small
	\begin{tabular}{l l }
            \toprule
		Type& Description\\
  \midrule
  \multirow{2}{*}{Factual} &Seeking specific, clear facts or evidence\\
  &\textit{Where is the capital of the United States?}\\
  \midrule
  \multirow{2}{*}{Analytical} &Seeking analysis for concepts, terms \\
  &\textit{Why is the earth warming? } \\
  \midrule
   \multirow{2}{*}{Comparative} & Seeking comparisons in different dimensions\\
   &\textit{What are the differences between A and B?}\\
   \midrule
    \multirow{2}{*}{Tutorial} &Seeking the steps to perform a task or process\\
    &\textit{What are the steps to install TensorFlow?}\\
		\bottomrule
	 \end{tabular}
	\caption{Definitions and examples of four types of queries.}
	\label{query_define}
\end{table}

\section{Preliminaries}
In this paper, we divide the whole process of RAG into four stages, including chunking, retrieval, reranking, and generation.
\textbf{Chunking} involves dividing the entire knowledge base into chunks according to chunk size with overlap between adjacent chunks.
\textbf{Retrieval} refers to converting both the query and chunks into numerical vectors using the embedding model and then selecting the top-K chunks as initial retrieved results based on the similarity between the query vector and the chunk vector.
\textbf{Reranking} refers to using the reranking model to understand the query and chunk to further rank the initial retrieved chunks and select the top-k as the final results.
\textbf{Generation} means leveraging LLMs to generate the response based on the query and final retrieved results.

\begin{figure}[t]
\centering
\includegraphics[width=\linewidth]{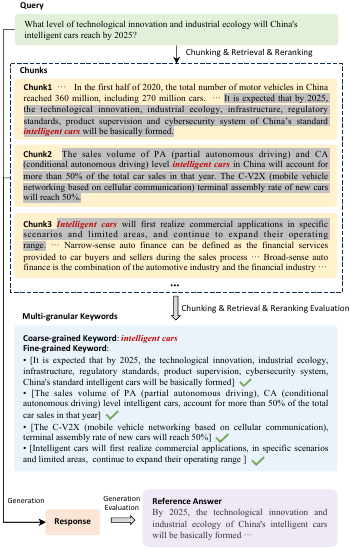} 
\caption{An example of the proposed CoFE-RAG framework. The red words denote coarse-grained keywords. The gray highlighted part is the corresponding content for fine-grained keywords.}
\label{framework}
\end{figure}

\begin{figure*}[t]
\centering
\includegraphics[width=\linewidth]{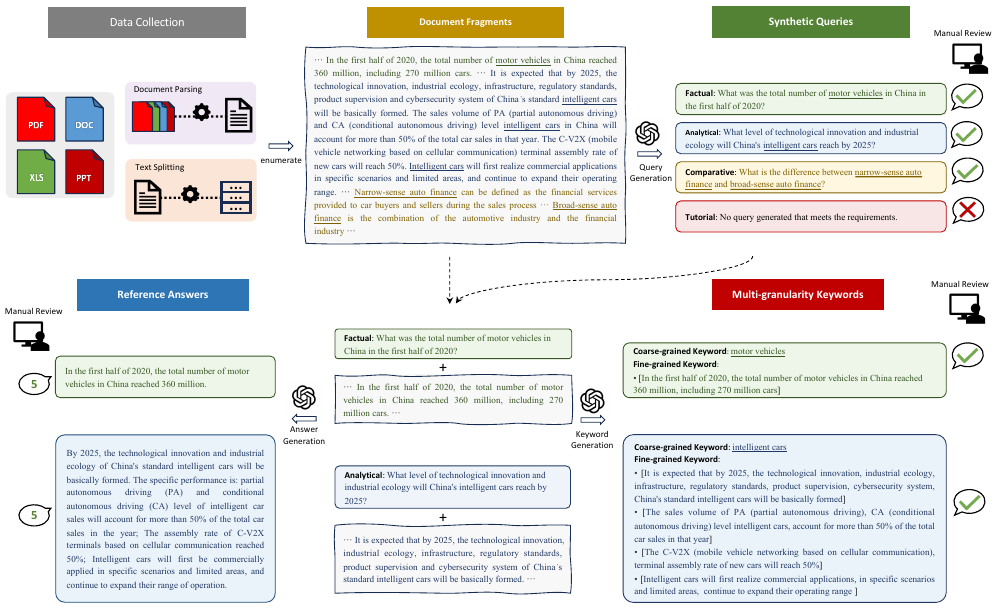} 
\caption{An example of the constructing process of query, multi-granularity keywords, and reference answers.}
\label{example}
\end{figure*}

\section{The CoFE-RAG Framework}
In this section, we demonstrate the proposed CoFE-RAG framework in detail, which aims to evaluate all phases of RAG systems containing chunking, retrieval, reranking, and generation. We introduce multi-granular keywords to facilitate a robust evaluation of chunking, retrieval, and reranking performance. 
The detailed process of the proposed CoFE-RAG framework is illustrated in Fig. \ref{framework}.

\subsection{Data Collection}
\subsubsection{Document Collection}
We collect a variety of documents from open-source websites, encompassing multiple formats such as PDF, DOC, PPT, and XLSX. These documents cover various industries, including finance, technology, medical care, commerce, Internet, etc. Their content includes industry reports, manuals, statistics, etc., providing a rich source of information suitable for evaluating RAG systems. The majority of the documents were created in recent years, with a considerable portion dating from this year (2024). This time frame surpasses the knowledge cutoff range of many widely used LLMs, such as GPT-4 \cite{DBLP:journals/corr/abs-2303-08774}. The distributions of documents across different formats are shown in Table \ref{doc_distribution}. 

\subsubsection{Document Parsing and Splitting} 
In the initial phase, we parse the documents to extract content suitable for processing by language models. 
Documents in PDF, PPT, and DOC formats are parsed by the LlamaIndex tool \cite{Liu_LlamaIndex_2022}, and the Pandas \cite{reback2020pandas} library is used to table content from XLSX documents. 
Then we split the content of each document into multiple fragments for subsequent data construction. 
To address the potential absence of title information in intermediate fragments, we employ GPT-4 to extract key information from the first fragment of each document. Such key information is then used as the title and appended to the beginning of each fragment.

\subsection{Data Construction}
The data construction process includes query generation, multi-granularity keywords generation, and reference answer generation, which is illustrated in Fig. \ref{example}. 
\subsubsection{Query Generation}
We define four distinct types of queries, including factual, analytical, comparative, and tutorial queries. Definitions for each query type are demonstrated in Table \ref{query_define}. 
We meticulously design prompts including task instruction, demonstration examples, and document fragment. 
For each document fragment, we employ GPT-4 to thoroughly comprehend the content and generate corresponding queries for all four types. 
It should be noted that if no applicable query can be generated for a specific query type that meets the requirements, the corresponding output will be \textit{It cannot be generated}.

We establish three essential criteria that a high-quality query must satisfy: (1) The query must be clear, precise, and free from grammatical errors, avoiding the use of ambiguous pronouns such as he, it, this, etc; (2) The query must align with the definition of its respective query type; (3) The query should be inferable from the information presented in the corresponding document fragment.
Then we employ well-trained annotators to assess the acceptability of each query. A query is deemed acceptable only if it fully complies with all the criteria.

\subsubsection{Multi-granularity Keywords Generation}
To address the issue of evaluating retrieval performance depending on golden chunks, we propose annotating multi-granularity keywords for each query instead. 
This approach eliminates the need for the labor-intensive process of re-labeling when the chunking strategy changes.

The multi-granularity keywords consist of coarse-grained and fine-grained keywords. Specifically, coarse-grained keywords are the most representative and relevant words extracted from the query and fragment, typically comprising one or two words that succinctly encapsulate the main topic. 
Fine-grained keywords are formulated as a set of lists, with each list corresponding to an information point extracted from the fragment. The elements of the list are specific spans of text taken directly from the original fragment, serving as reference points for answering the query.

For example in Fig. \ref{example}, for the analytical query \textit{What level of technological innovation and industrial ecology will China's intelligent cars reach by 2025?}, we first extract the coarse-grained keywords \textit{intelligent cars}. To adequately address this query, we identify four distinct information points from the document fragment, each corresponding to a separate list.

Like the query generation process, we utilize GPT-4 to generate coarse-grained keywords and fine-grained keywords with carefully designed prompt containing task instruction, demonstration examples, query, and document fragment.
If no suitable coarse-grained or fine-grained keyword can be generated that meets the requirements, the resulting output list will be left blank.

To ensure quality, well-trained annotators are then employed to evaluate the acceptability of all coarse-grained keywords and calculate the acceptance rate for fine-grained keywords. We retain only those examples where all coarse-grained keywords are accepted and the acceptance rate for fine-grained keywords exceeds 80\%. 
The acceptance rate means how many correct lists are recalled, with a list considered correct only when each of its elements is correct.
This meticulous process ensures the reliability and quality of the multi-granularity keywords, facilitating a robust and nuanced evaluation of retrieval-augmented generation systems.

\subsubsection{Reference Answer Generation}
We provide a reference answer for each query to serve as a benchmark for evaluating the generation performance of RAG systems.
Similarly, we employ GPT-4 to generate reference answers with meticulously crafted prompt.
To ensure the quality of these reference answers, we ask annotators to evaluate them based on five criteria: fluency, accuracy, relevance, readability, and practicality. Each answer is scored on a scale from 1 to 5 points. We then filter out samples with answer scores below 4 points to maintain a high standard of quality. This stringent filtering process ensures that only high-quality reference answers are retained for evaluations.

\begin{table}[!t]
	\centering
   \small
	\begin{tabular}{l ccc}
            \toprule
            Query Type&Raw&Final&Accept Rate(\%)\\
            \midrule
	Factual &1786&1340&75.0\\
Analytical &1489&746&50.1\\
Comparative &903&498&55.1\\
Tutorial 	&513&242&47.2\\
\midrule
Total&4691&2826&60.2\\
   \bottomrule
	 \end{tabular}
	\caption{The distribution of query types, where \textit{Raw} and \textit{Final} represent the number of queries before and after manual review.}
	\label{query_type}
\end{table}

\subsection{Data Statistics}
After three generation steps, we obtain examples consisting of queries, coarse-grained keywords, fine-grained keywords, and reference answers. The generated data went through rigorous human review to ensure high quality.
For synthetic queries, we observe that 92.2\% of them are accepted by human annotators.
For synthetic multi-granularity keywords, 87.3\% of them are accepted by human annotators (The coarse-grained keywords are accepted and the acceptance rate of fine-grained keywords is larger than 80\%).
For generated reference answers,  74.8\% of them are accepted by human annotators. 
Thus, the overall acceptance rate after manual review is 60.2\%.

The distribution of query types is detailed in Table \ref{query_type}.
Among all types of queries, factual queries account for the largest proportion.  
This is attributable to the higher generation rate and the larger proportion of factual queries meeting the filtering criteria.
Conversely, tutorial queries have the smallest proportion, largely due to the original documents containing limited tutorial information, which in turn results in fewer queries of this type.

\subsection{Evaluation Metrics}
We utilize a series of evaluation metrics to assess all stages of RAG systems. 

\subsubsection{Chunking \& Retrieval \& Reranking Evaluation}

The proposed CoFE-RAG aims to evaluate the chunking, retrieval, and reranking quality based on multi-granularity keywords rather than golden chunks. 
For the top-K retrieval chunks, we regard coarse-grained keywords as a loose constraint and filter out the results that do not contain any coarse-grained keywords. This step ensures that only contextually relevant chunks are considered for further evaluation. After filtering, we concatenate the remaining chunks and use two metrics to evaluate the results, including Recall and Accuracy. 

Specifically, Recall evaluates how many fine-grained keyword lists are correctly recalled from all the annotated fine-grained keyword lists of the whole dataset.
Accuracy reflects the ratio of completely correct retrieved results among all examples. A result is considered completely correct when all fine-grained keyword lists of an example are correctly recalled.



\subsubsection{Generation Evaluation}
We utilize various metrics to evaluate the quality of generated response, including BLEU \cite{DBLP:conf/acl/PapineniRWZ02}, Rouge-L \cite{lin2004rouge}, Faithfulness, Relevance, and Correctness.

Specifically, BLEU measures the similarity between the generated response and the reference answer by calculating the n-gram exact match between them.
Rouge-L measures the similarity between the generated response and the reference answer by the Longest Common Subsequence (LCS), focusing on order and coverage.
Faithfulness, Relevance, and Correctness are calculated by the built-in evaluator of LlamaIndex, which uses GPT-4 to automatically evaluate via in-context learning.
Faithfulness evaluates whether a generated response is faithful to the retrieved context.
Relevance evaluates the relevancy of retrieved context and generated response to a query. 
Correctness evaluates the correctness of the system. This evaluator can output a score between 1 and 5 based on the query, generated response, and reference answer, where 1 is the worst and 5 is the best, as well as the reason for the score.
Score represents the average correctness score of all examples. 
Pass is defined as the ratio of examples whose score is greater than or equal to 4.

\begin{table*}[!t]
	\centering
	\resizebox{\textwidth}{!}{
	\begin{tabular}{l cccccccccc}
            \toprule
            \multirow{2}{*}{Embedding}&\multicolumn{2}{c}{Factual}&\multicolumn{2}{c}{Analytical}&\multicolumn{2}{c}{Comparative}&\multicolumn{2}{c}{Tutorial}&\multicolumn{2}{c}{Overall} \\
            \cmidrule(r){2-3} \cmidrule(r){4-5} \cmidrule(r){6-7} \cmidrule(r){8-9} \cmidrule(r){10-11}
            &Recall&Accuracy&Recall&Accuracy&Recall&Accuracy&Recall&Accuracy&Recall&Accuracy\\
            \midrule
  
 text-embedding-ada-002&0.6288 &0.5833 &0.6027 &0.5691 &0.6067 &0.5594 &0.5772 &0.4938 &0.6080 &0.5669\\
  text-embedding-3-large&0.6763 &0.6385 &0.6603 &0.6067 &0.6471 &0.6056 &0.6131 &0.5477 &0.6565 &0.6157\\
stella-large&0.7525 &0.6968 &0.7091 &0.6443 &0.6700 &0.6298 &0.7006 &0.6224 &0.7142 &0.6638\\
m3e-large&0.6915 &0.6303 &0.6496 &0.5732 &0.6096 &0.5493 &0.6608 &0.5726 &0.6566 &0.5952\\
piccolo-large&0.7442 &0.6893 &0.6827 &0.6255 &0.6630 &0.6237 &\textbf{0.7070} &0.6100 &0.7011 &0.6532\\
gte-large &0.6898 &0.6378 &0.6537 &0.5933 &0.6348 &0.5875 &0.6752 &0.5892 &0.6641 &0.6122\\
bge-base&0.7470 &0.6871 &0.7108 &0.6443 &0.6717 &0.6258 &0.6855 &0.6141 &0.7114 &0.6578\\
bge-large&\textbf{0.7612} &\textbf{0.7028} &\textbf{0.7124} &\textbf{0.6591} &\textbf{0.6735} &\textbf{0.6378} &0.7030 &\textbf{0.6224} &\textbf{0.7190} &\textbf{0.6720}\\
		\bottomrule
	 \end{tabular}}
	\caption{Retrieval performance of baselines on the dataset.}
	\label{retrieval_result}
\end{table*}

\begin{table*}[!t]
	\centering
	\resizebox{\textwidth}{!}{
	\begin{tabular}{l cccccccccc}
            \toprule
            \multirow{2}{*}{Reranking} &\multicolumn{2}{c}{Factual}&\multicolumn{2}{c}{Analytical}&\multicolumn{2}{c}{Comparative}&\multicolumn{2}{c}{Tutorial}&\multicolumn{2}{c}{Overall} \\
            \cmidrule(r){2-3} \cmidrule(r){4-5} \cmidrule(r){6-7} \cmidrule(r){8-9} \cmidrule(r){10-11}
            &Recall&Accuracy&Recall&Accuracy&Recall&Accuracy&Recall&Accuracy&Recall&Accuracy\\
            \midrule
  jina-reranker-v2-base&0.7175 &0.6699 &0.6559 &0.5987 &0.6096 &0.5714 &0.6330 &0.5560 &0.6633 &0.6231\\
  bce-reranker-base&0.7251 &0.6721 &\textbf{0.6678} &0.6040 &0.6102 &0.5775 &0.6457 &0.5613 &\textbf{0.6719} &0.6270\\
  bge-reranker-base&0.7220 &0.6714 &0.6537 &0.5919 &\textbf{0.6120 }&0.5782 &0.6417 &0.5602 &0.6654 &0.6238\\
  bge-reranker-large&\textbf{0.7262} &\textbf{0.6759} &0.6625 &\textbf{0.6067 }&0.6114 &\textbf{0.5795} &\textbf{0.6529 }&\textbf{0.5685} &0.6714 &\textbf{0.6306}\\
		\bottomrule
	 \end{tabular}}
	\caption{Reranking performance of baselines on the dataset.}
	\label{rerank_result}
\end{table*}

\begin{table*}[!t]
	\centering
	\setlength\tabcolsep{10pt}
        \small
	\begin{tabular}{l cccccc}
            \toprule
            \multirow{2}{*}{LLM} & \multirow{2}{*}{BLEU}&\multirow{2}{*}{Rouge-L}&\multirow{2}{*}{Faithfulness}&\multirow{2}{*}{Relevance}&\multicolumn{2}{c}{Correctness}\\
            \cmidrule(r){6-7} 
            &&&&&Pass&Score \\
            \midrule
Qwen2-0.5B&0.1650&0.3126&0.7367&0.7824&0.3443&2.7093\\
Qwen2-1.5B&0.1437&0.3022&0.7385&0.7785&0.3439&2.9338\\
Qwen2-7B&0.2649&0.4925&0.8372&0.9253&0.6348&3.7699\\
Llama2-7B&0.2323&0.3345&0.8461&0.7611&0.3808&3.1175\\
ChatGLM3-6B&0.2662&0.4100&0.8659&0.8255&0.5180&3.3942\\
Claude-2.1&0.2141&0.4060&0.8742&0.9018&0.5612&3.3349\\
Claude-3-Opus&0.2623&0.5209&0.8846&\textbf{0.9565}&0.6684&3.8613\\
GPT-3.5-Turbo&0.2934&0.4215&\textbf{0.9222}&0.9176&0.5690&3.5290\\
GPT-4o&\textbf{0.4565}&\textbf{0.5519}&0.8977&0.9441&\textbf{0.7389}&\textbf{4.0777}\\
  \bottomrule
	 \end{tabular}
	\caption{Generation performance of baselines on the dataset.}
	\label{generation_result}
\end{table*}

\begin{figure*}[htbp]
    \centering
    \begin{subfigure}{0.32\textwidth} 
        \centering
        \includegraphics[width=\linewidth]{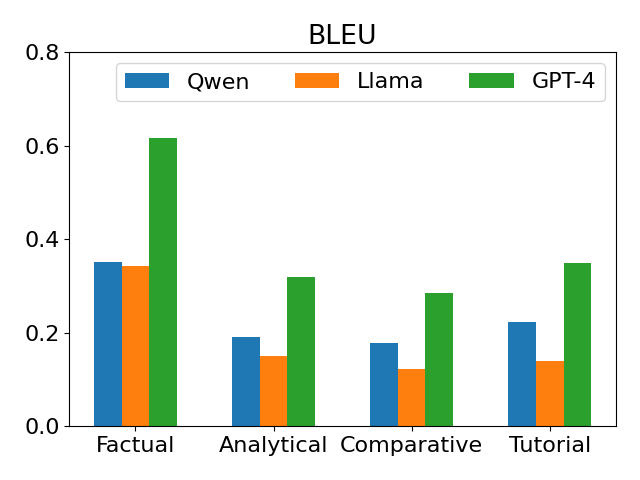}
        \label{fig:sub1}
    \end{subfigure}%
    \hfill 
    \begin{subfigure}{0.32\textwidth}
        \centering
        \includegraphics[width=\linewidth]{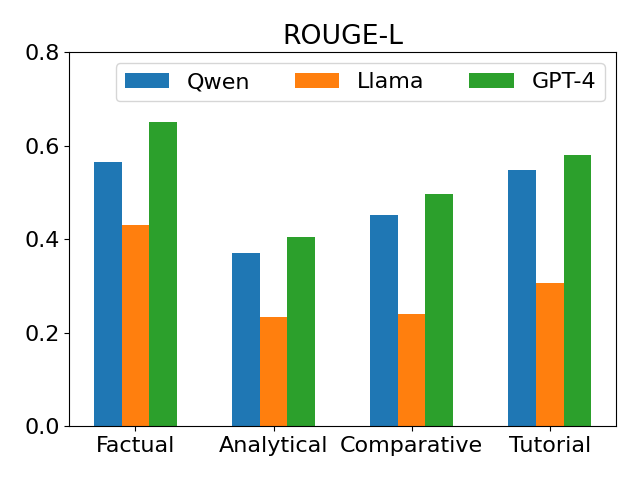}
        \label{fig:sub2}
    \end{subfigure}%
    \hfill 
    \begin{subfigure}{0.32\textwidth}
        \centering
        \includegraphics[width=\linewidth]{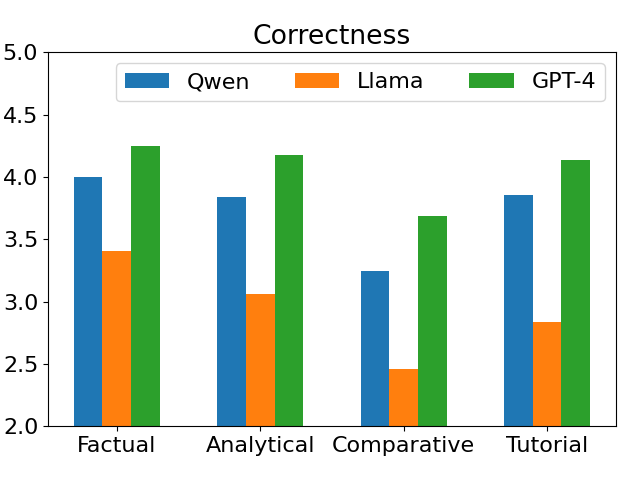}
        \label{fig:sub3}
    \end{subfigure}
    \caption{BLEU, Rouge-L, and Correctness score over different query types.}
    \label{fig_gen}
\end{figure*}

\begin{figure}[t]
\centering
\includegraphics[width=7cm]{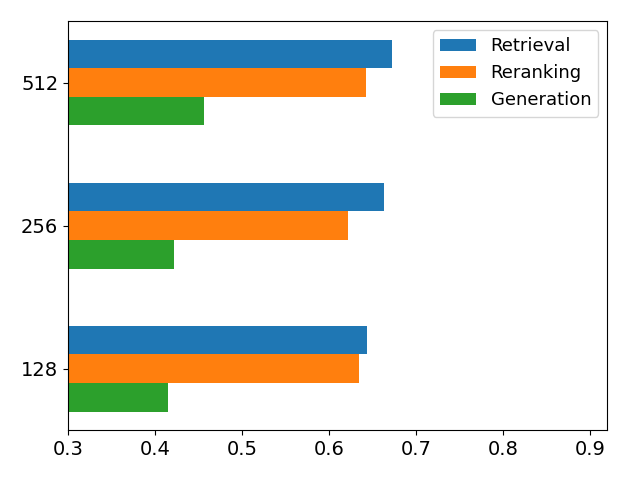} 
\caption{Experimental results with different chunk size. The retrieval and reranking phases are evaluated by Accuracy, while the generation stage is assessed by BLEU.}
\label{chunksize}
\end{figure}

\section{Experiments}
The proposed dataset can be used as a benchmark for evaluating RAG systems in more diverse data scenarios.
In this section, we conduct experiments to demonstrate the effect of retrieval, reranking, generation, and chunking, respectively.

\subsection{Effect of Retrieval}
We first split all documents into chunks with a size of 512 tokens, with an overlap of 100 tokens between two adjacent chunks. We use the top 30 chunks as initial retrieved results to evaluate retrieval performance. We choose a variety of embedding models, include text-embedding-ada-002 and text-embedding-3-large by OpenAI \cite{textada2}, stella-large-zh-v2 \cite{stella_mbedding_2023}, m3e-large \cite{m3e}, piccolo-large-zh-v2 \cite{DBLP:journals/corr/abs-2405-06932}, gte-large-zh \cite{DBLP:journals/corr/abs-2308-03281}, bge-base-zh-v1.5, and bge-large-zh-v1.5 \cite{DBLP:journals/corr/abs-2309-07597}. 

The experimental results for different embedding models are shown in Table \ref{retrieval_result}. 
We observed that the bge-large model outperforms others in terms of Recall and Accuracy across all types of queries and overall performance. This indicates that the model has a strong ability to capture the semantic relationship between queries and their context.
Among all embedding models, factual queries generally perform better than analytical, comparative, and tutorial queries. This may be because the relevant context for factual queries is usually contained within a single chunk, making it easier to retrieve. In contrast, other types of queries are more complex, with relevant context potentially spread across multiple chunks, making retrieval more challenging.
Additionally, existing retrieval models generally suffer from poor performance, highlighting the ongoing challenge of searching relevant chunks that accurately match the query.

\subsection{Effect of Reranking }

We rerank the initial retrieved results and select the top 4 chunks to assess the reranking performance. 
To evaluate the reranking methods, we use the chunks retrieved by bge-large-zh-v1.5 and conduct experiments with various reranking models, including jina-reranker-v2-base-multilingual \cite{gunther2023jina}, bce-reranker-base \cite{youdao_bcembedding_2023}, bge-reranker-base, and bge-reranker-large \cite{DBLP:journals/corr/abs-2309-07597}.

The experimental results with different reranking models are reported in Table \ref{rerank_result}. We can observe that bge-reranker-large stands out with the best performance. 
Additionally, using the reranked top 4 results proves less effective compared to utilizing all retrieved results. 
This indicates that the current reranking methods are still not performing well and may miss some relevant chunks. 
After the retrieval and reranking phases, the performance of factual queries still outperforms the other three queries, which further demonstrates our analysis.

\subsection{Effect of Generation}



The generation stage has a great impact on the RAG system, as different LLMs vary in their ability to integrate queries and retrieved chunks to generate responses. We feed the query and top 4 chunks reranked by bge-reranker-large into various LLMs for evaluation. 
Our experiments encompass a diverse array of LLMs, including GPT-4o, GPT-3.5-Turbo \cite{DBLP:journals/corr/abs-2303-08774}, Claude-2.1, Claude-3-Opus \cite{anthropic2023claude2}, Qwen2 \cite{qwen2}, Llama2 \cite{DBLP:journals/corr/abs-2302-13971}, and ChatGLM3 \cite{du2022glm}. 

The generation performance with different LLMs is reported in Table \ref{generation_result}.
We observed that GPT-4 achieved the best results across various LLMs, significantly outperforming other models.
Models with larger parameters, such as GPT-4 and Claude-3 generally perform better than models with smaller parameters, such as Qwen-7B, Llama-7B. This may be because models with larger parameters have stronger reasoning and generalization capabilities, reduce the risk of hallucinations, and can handle more complex tasks.
Qwen2-7B performs the best among Qwen2-7B, Llama2-7B, and ChatGLM-6B, demonstrating its ability to generate accurate and reliable answers in the RAG system.

To provide a more detailed comparison, we present the BLEU, Rouge-L, and Correctness scores for Qwen2-7B, Llama2-7B, and GPT-4 across different query types in Figure \ref{fig_gen}.
We can observe that the performance on factual queries generally outperforms the other query types. 
This observation highlights the complexity and challenging nature of analytical, comparative, and tutorial queries, suggesting that further efforts are required to enhance performance on these more intricate query types.

\subsection{Effect of Chunking}
To demonstrate the effect of chunking, we conduct experiments with chunk sizes of 128, 256, and 512 tokens, respectively. 
The corresponding overlap sizes are set to 25, 50, and 100 tokens, and the final number of chunks after reranking is set to 16, 8, and 4, respectively. 
For these experiments, we employed the bge-large-zh-v1.5 model for retrieval, the bge-reranker-large model for reranking, and GPT-4o for generation.
The performance with different chunk sizes is illustrated in Fig. \ref{chunksize}.
We can observe that using a size of 512 can achieve better retrieval, reranking, and generation performance. This indicates that larger chunks are more effective at preserving the original information from the document, thereby benefiting the ability of the system to address complex queries.

\section{Conclusion}
In this paper, we present the CoFE-RAG framework to facilitate thorough evaluation across the entire RAG pipeline. We introduce multi-granularity keywords to assess the retrieved context instead of relying on the annotation of golden chunks, which can effectively evaluate chunking, retrieval, and reranking performance particularly when the chunking strategy changes.
Moreover, we release a holistic benchmark dataset tailored for diverse data scenarios covering a wide range of document formats and query types.
The experimental results indicate that while there have been significant advancements, current methods still have substantial room for improvement, particularly in handling complex query types and diverse knowledge sources.


\bibliography{aaai24}

\clearpage

\appendix
\section{Experimental Results on English Queries}

\begin{table}[h]
	\centering
	\begin{tabular}{l cccc}
            \toprule
            \multirow{2}{*}{Query Type}&\multicolumn{2}{c}{English}&\multicolumn{2}{c}{Chinese}\\
            &Count&Ratio&Count&Ratio\\
            \midrule
	Factual &364&36.3\%&1340&47.4\%\\
Analytical &260&25.9\%&746&26.4\%\\
Comparative &226&22.5\%&498&17.6\%\\
Tutorial 	&153&15.3\%&242&8.6\%\\
\midrule
Total&1003&-&2826&-\\
   \bottomrule
	 \end{tabular}
	\caption{The distribution of query types on English and Chinese queries.}
	\label{query_type_en}
\end{table}

The proposed dataset contains queries in both Chinese and English languages. The distributions of the query types are shown in Table \ref{query_type_en}. In the main body of the paper, we mainly conduct experiments and analysis on Chinese queries. 
In the appendix, we present benchmark experimental results on English queries with the same document base.

In the implementations, we employ bge-large-en-v1.5 as the embedding model, bge-reranker-large as the reranking model, and GPT-4o as LLMs for generation. We use a chunk size of 512 tokens with an overlap of 100 tokens. We first retrieve the top 30 chunks using the embedding model. Then we rank these chunks using the reranking model and select the top 4 chunks for generation. The experimental results are demonstrated in Table \ref{english_result} and Table \ref{generation_result_en}.

\begin{table*}[!t]
	\centering
	\resizebox{\textwidth}{!}{
	\begin{tabular}{l cccccccccc}
            \toprule
            \multirow{2}{*}{} &\multicolumn{2}{c}{Factual}&\multicolumn{2}{c}{Analytical}&\multicolumn{2}{c}{Comparative}&\multicolumn{2}{c}{Tutorial}&\multicolumn{2}{c}{Overall} \\
            \cmidrule(r){2-3} \cmidrule(r){4-5} \cmidrule(r){6-7} \cmidrule(r){8-9} \cmidrule(r){10-11}
            &Recall&Accuracy&Recall&Accuracy&Recall&Accuracy&Recall&Accuracy&Recall&Accuracy\\
            \midrule
            Retrieval&0.7648 &0.7308 &0.6661 &0.4077 &0.6348 &0.4425 &0.6519 &0.5098 &0.6765 &0.5484\\
  Reranking&0.7402 &0.7198 &0.5931 &0.3731 &0.5703 &0.4159 &0.5711 &0.4771 &0.6129 &0.5244\\
		\bottomrule
	 \end{tabular}}
	\caption{Retrieval and reranking performance of baselines on the English queries.}
	\label{english_result}
\end{table*}

\begin{table*}[!t]
	\centering
	\begin{tabular}{l cccccc}
            \toprule
            \multirow{2}{*}{} & \multirow{2}{*}{BLEU}&\multirow{2}{*}{Rouge-L}&\multirow{2}{*}{Faithfulness}&\multirow{2}{*}{Relevance}&\multicolumn{2}{c}{Correctness}\\
            \cmidrule(r){6-7} 
            &&&&&Pass&Score \\
            \midrule
Generation&0.5016&0.5666&0.9332&0.9671&0.7358&4.0304\\
  \bottomrule
	 \end{tabular}
	\caption{Generation performance of baselines on the English queries.}
	\label{generation_result_en}
\end{table*}


\section{An Example of the Dataset}
We leverage coarse-grained keywords and fine-grained keywords to evaluate the chunking, retrieval, and reranking stages, while using reference answer to assess the generated response.
Here we present an example in json format:

\noindent \{\\
\noindent "\textbf{query type}": "Analytical",\\
         \noindent   "\textbf{query}": "What are the main responsibilities of a Program Support Assistant (Office Automation) in the Research and Development Service?",\\
            "\textbf{coarse-grained keywords}": [\\
                "Program Support Assistant"\\
            ],\\
            "\textbf{fine-grained keywords}": [
            
       \noindent         [
                    "prepare proposal forms",
                    "monitors each research proposal",
                    "reviews each proposal for conformance with policies, regulations, format, completeness and accuracy"
                ],
                
      \noindent          [
                    "Establishes agendas",
                    "prepares or coordinates preparation of monthly packages",
                    "schedules meetings",
                    "takes notes of the proceedings",
                    "converts the notes into formal minutes",
                    "prepares these minutes for the chairperson’s review and signature",
                    "distributes them"
                ],
                
      \noindent          [
                    "Monitors each project",
                    "notifies the investigators of the need for reports",
                    "incorporates these reports into the project package",
                    "initiates and maintains a file of all abstracts and publications"
                ],
                
      \noindent          [
                    "Manages all project data activities utilizing the PROMISE system"
                ],
                
    \noindent            [
                    "Acts on requests for information",
                    "answers questions related to procedure",
                    "providing copies of application forms and instructions",
                    "reviewing with them the major requirements"
                ],
                
   \noindent             [
                    "Establishes and maintains a personal calendar",
                    "notation of action dates for initial submission of material or follow-up"
                ],
                
    \noindent            [
                    "Performs other duties as assigned"
                ]
                
    \noindent        ],\\
            "\textbf{reference answer}": \\
            "The main responsibilities of a Program Support Assistant (Office Automation) in the Research and Development Service include:\\
   \noindent         1. Assisting investigators with proposal preparation and monitoring research proposals from submission to completion, ensuring conformance with policies, regulations, and accuracy.
            
   \noindent         2. Establishing agendas and scheduling meetings for the Research and Development Committee, Human Studies Subcommittee, and Animal Studies Subcommittee.
            
    \noindent        3. Preparing and coordinating monthly packages for committee members, taking meeting notes, converting them into formal minutes, and distributing them.
            
    \noindent        4. Monitoring project reports (initial, progress, final), notifying investigators of required reports, maintaining files of abstracts and publications, and preparing them for distribution.
            
    \noindent        5. Managing project data activities using the PROMISE system.
            
     \noindent       6. Responding to information requests from potential grant applicants, providing application forms and instructions, and reviewing major requirements with them.
            
     \noindent       7. Maintaining a personal calendar for various grant deadlines and action dates.
            
     \noindent       8. Performing other assigned duties."\\
        \}

\end{document}